\documentclass{article}
\usepackage{spconf}

\usepackage{textcomp}

\usepackage{amsmath}
\usepackage{amssymb}
\usepackage{booktabs}
\usepackage{cite}
\usepackage{color}
\usepackage{enumitem}
\usepackage{gensymb}
\usepackage{graphicx}
\usepackage{latexsym}
\usepackage{multirow}
\usepackage{nonfloat}
\usepackage[all]{nowidow}
\usepackage{pifont}
\usepackage[caption=false,farskip=3pt,captionskip=3pt]{subfig}
\usepackage{xfrac}

\usepackage{hyperref}
\hypersetup{breaklinks=true,colorlinks=true}

\title{Reflectance-Oriented Probabilistic Equalization\\
for Image Enhancement}
\name{Xiaomeng Wu$^{\dagger}$, Yongqing Sun$^{\ddagger}$, Akisato Kimura$^{\dagger}$, and Kunio Kashino$^{\dagger}$}
\address{$^{\dagger}$ Communication Science Laboratories, NTT Corporation, Japan \\
$^{\ddagger}$ Media Intelligence Laboratories, NTT Corporation, Japan}

\begin{document}
\hyphenation{Re-flectance Re-flectances Hy-per-pa-ra-me-ter Hy-per-pa-ra-me-ters De-nois-ing Trade-off Dis-crim-i-na-tive Adap-tive-ly In-ter-pretable Gray-scale Ob-ject-ness Color-Checker Mo-not-o-nic-i-ty Data-set Data-sets Con-vo-lu-tion-al It-er-a-tive-ly}
\ninept
\maketitle

\begin{abstract}
Despite recent advances in image enhancement, it remains difficult for existing approaches to adaptively improve the brightness and contrast for both low-light and normal-light images. To solve this problem, we propose a novel 2D histogram equalization approach. It assumes intensity occurrence and co-occurrence to be dependent on each other and derives the distribution of intensity occurrence (1D histogram) by marginalizing over the distribution of intensity co-occurrence (2D histogram). This scheme improves global contrast more effectively and reduces noise amplification. The 2D histogram is defined by incorporating the local pixel value differences in image reflectance into the density estimation to alleviate the adverse effects of dark lighting conditions. Over 500 images were used for evaluation, demonstrating the superiority of our approach over existing studies. It can sufficiently improve the brightness of low-light images while avoiding over-enhancement in normal-light images.
\end{abstract}

\begin{keywords}
2D histogram equalization, reflectance, Retinex model, contrast enhancement, image enhancement
\end{keywords}

\section{Introduction}
\label{s:introduction}

Image enhancement aims to enhance image contrast and reveal hidden image details. With the rapid development of digital imaging devices, the number of images out there and the demand for image enhancement has increased significantly. Commercial raster graphics editors require image editing expertise or considerable manual effort to produce satisfactory image enhancement. Therefore, it is essential to develop an automated image enhancement technique that adapts to different input lighting conditions. Existing approaches to image enhancement can be classified into model-based approaches and learning-based ones. We focus on model-based approaches, as they are more interpretable and do not need labeled training data.

In model-based approaches, histogram equalization (HE) has received the most attention. It derives an intensity mapping function such that the entropy of the distribution of output intensities is maximized. However, HE extends the contrast between intensities with large populations to a wider range, even if it is not semantically important. This issue has been addressed by incorporating spatial information into density estimation~\cite{AriciDA09,CelikT11,Celik12,EilertsenMU15,SuJ17,WuLHK17,WuKK20}. For example, 2DHE~\cite{CelikT11,Celik12} equalizes the 2D histogram of intensity co-occurrence so that the contrast between frequently co-occurring intensities is enhanced to a greater extent. CACHE~\cite{WuLHK17} incorporates image gradients into histogram construction to avoid the excessive enhancement of trivial background. However, such spatial information is not discriminative enough, especially for low-light image areas. Their equalization schemes~\cite{CelikT11,Celik12,WuLHK17} also overemphasize the importance of the frequently co-occurring intensities, tending to cause precipitous brightness fluctuation in very dark or very bright image areas.

Another direction~\cite{JobsonRW97,JobsonRW97a,Ramponi98,DurandD02,FarbmanFLS08,WangZHL13,FuZHZD16,GuoLL17,WangL18,RenYLL19} is based on the Retinex model. It takes an image as a combination of illumination and reflectance components, which capture global brightness and sharp image details, respectively. Some studies~\cite{JobsonRW97,JobsonRW97a,Ramponi98,DurandD02,FarbmanFLS08} assumed reflectance to be the desired enhancement output and obtained it by estimating and removing illumination. However, this strategy sometimes leads to excessively enhanced brightness. In other studies, LIME~\cite{GuoLL17} assumes that the gamma correction of the reflectance is the ideal form of low-light image enhancement. In NPE~\cite{WangZHL13} and NPIE~\cite{WangL18}, the illumination is enhanced with HE and recombined with the reflectance to reconstruct the enhanced image. Ren et al.~\cite{RenYLL19} found that the illumination can be leveraged as the exposure ratio of a camera response function (CRF), and proposed a novel CRF-based image enhancement approach called LECARM. These approaches are valid for discovering dark image details. However, it is not easy for them to find a solution optimized for both low-light and normal-light images; they tend to overly amplify the brightness and saturation of normal-light images.

Here, we propose a novel 2DHE approach known as reflectance-oriented probabilistic equalization (ROPE), which allows for adaptive regulation of global brightness. ROPE assumes intensity occurrence and co-occurrence to be dependent and derives the distribution of intensity occurrence (1D histogram) by marginalizing over the distribution of intensity co-occurrence (2D histogram). This scheme builds a novel bridge between 1D and 2D histograms. Compared to related approaches such as CVC~\cite{CelikT11} (2DHE) and CACHE~\cite{WuLHK17,WuKK20} (1DHE), ROPE provides more adequate contrast enhancement and less noise amplification. Inspired by RG-CACHE~\cite{WuKK20}, we define a novel 2D histogram by incorporating the local pixel value differences in image reflectance into the density estimation to alleviate the adverse effects of dark lighting conditions. Experiments show that ROPE outperforms state-of-the-art image enhancement approaches from both qualitative and quantitative perspectives.

\section{Proposed Approach}
\label{s:approach}

\subsection{Preliminaries}
\label{subs:preliminaries}

Given a color image $\mathbf{C}_\mathrm{in}$, its grayscale image $\mathbf{A}_\mathrm{in}$ is defined as the max of its RGB components and is equal to its value channel in HSV space~\cite{Smith78}. Let $\mathbf{A}_\mathrm{out}$ be the enhanced image of $\mathbf{A}_\mathrm{in}$ in ROPE. Let $\circ$ and $\oslash$ denote element-wise manipulation and division, respectively. The final output $\mathbf{C}_\mathrm{out}$ is computed as
\begin{equation}
\mathbf{C}_\mathrm{out}=(\mathbf{C}_\mathrm{in}\oslash\mathbf{A}_\mathrm{in})\circ\mathbf{A}_\mathrm{out}.
\label{e:color}
\end{equation}
Let $\mathbf{A}_\mathrm{in}=\{a(q)\}$ and $\mathbf{A}_\mathrm{out}=\{\hat{a}(q)\}$, where $a(q)\in[0,K)$ and $\hat{a}(q)\in[0,K)$ are the intensities of the pixels $q$ in the input and output, respectively. $K$ is the total number of possible intensity values (typically 256). Our goal is to find an intensity mapping function $T$ of the form $\hat{a}(q)=T(a(q))$ to produce the enhanced image.

Let $o_k$ be the event of $a(q)=k$ (occurrence), where $a(q)\in\mathbf{A}_\mathrm{in}$ and $k\in[0,K)$ is an intensity value. Let $p(o_k)$ be its probability. The 1D histogram of $\mathbf{A}_\mathrm{in}$ can then be expressed by $\{p(o_k)\}$. Let $P(o_k)$ be the cumulative distribution function of $p(o_k)$. In HE, the intensity mapping function $T(\cdot)$ is given by $T(k)=KP(o_k)-1$. The problem is how to properly define $p(o_k)$.

\subsection{Modeling of 1D Histogram as Marginal Probability}
\label{subs:se}

In ROPE, we define $p(o_k)$ based on the 2D histogram of $\mathbf{A}_\mathrm{in}$. Let $c_{i,j}$ be the event of $a(q)=i$ and $a(q')=j$ (co-occurrence), where $q'\in\mathcal{N}(q)$ and $\mathcal{N}(q)$ is the set of coordinates in the local window centered on the pixel $q$. Let $p(c_{i,j})$ be its probability. Then the 2D histogram can be written as $\{p(c_{i,j})\}$ with $i,j\in[0,K)$ and $i<j$. The construction of this 2D histogram is discussed in Section~\ref{subs:rope}.

\begin{figure}[!t]
\centering
\subfloat[CVC~\cite{CelikT11}]{
\includegraphics[scale=.2]{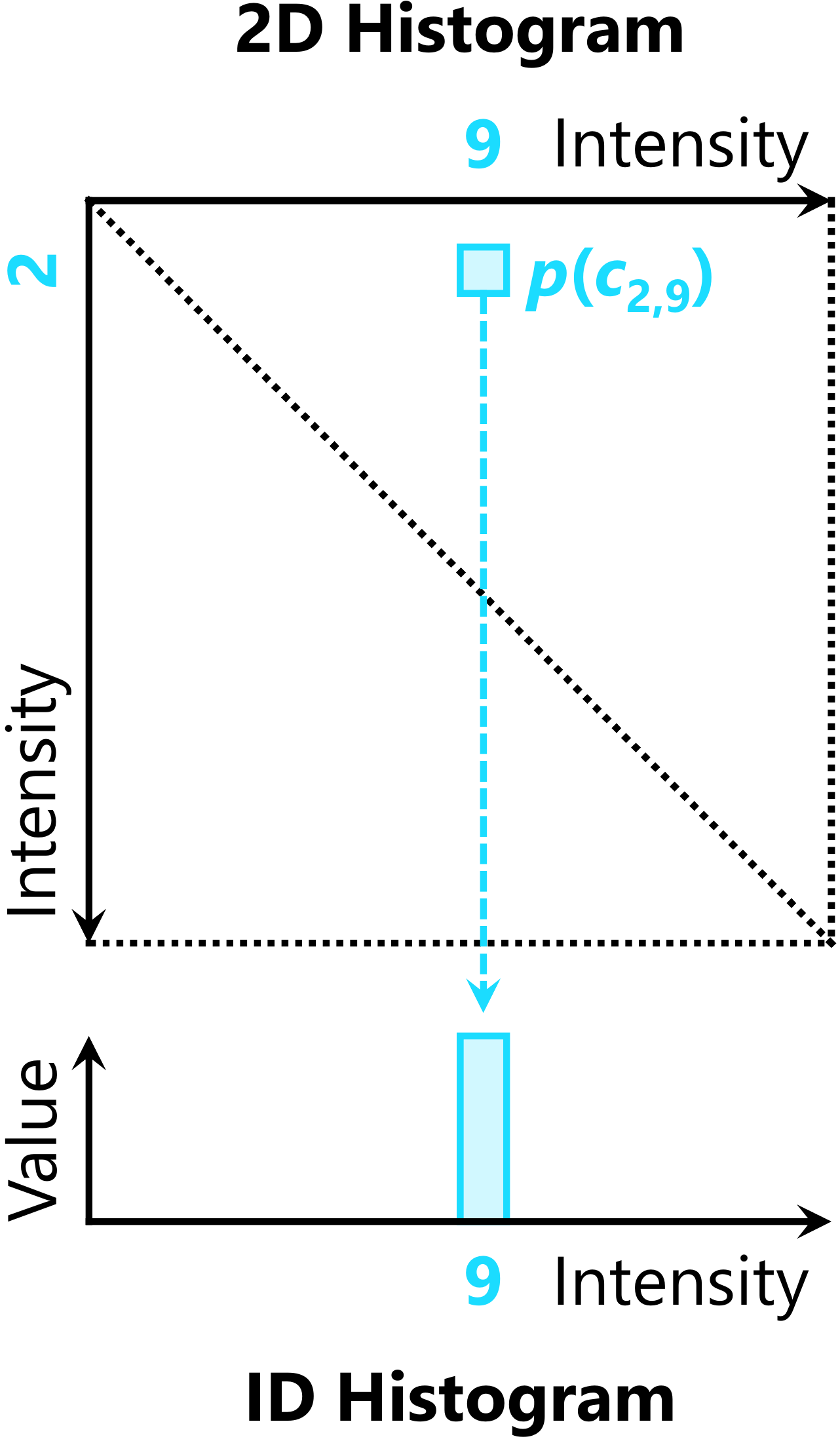}
\label{subf:cvc}}
\hfill
\subfloat[CACHE~\cite{WuLHK17,WuKK20}]{
\includegraphics[scale=.2]{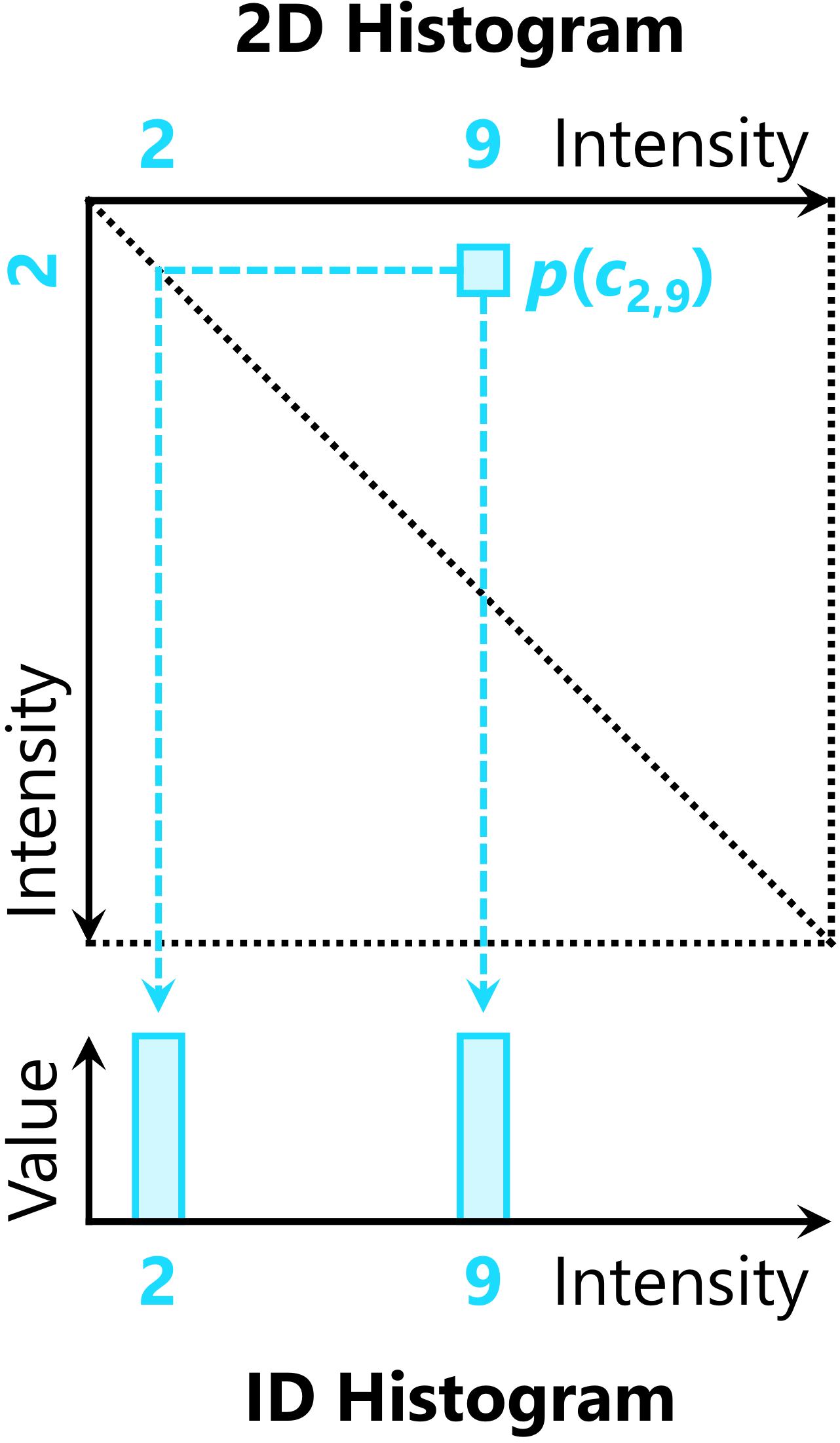}
\label{subf:cache}}
\hfill
\subfloat[ROPE]{
\includegraphics[scale=.2]{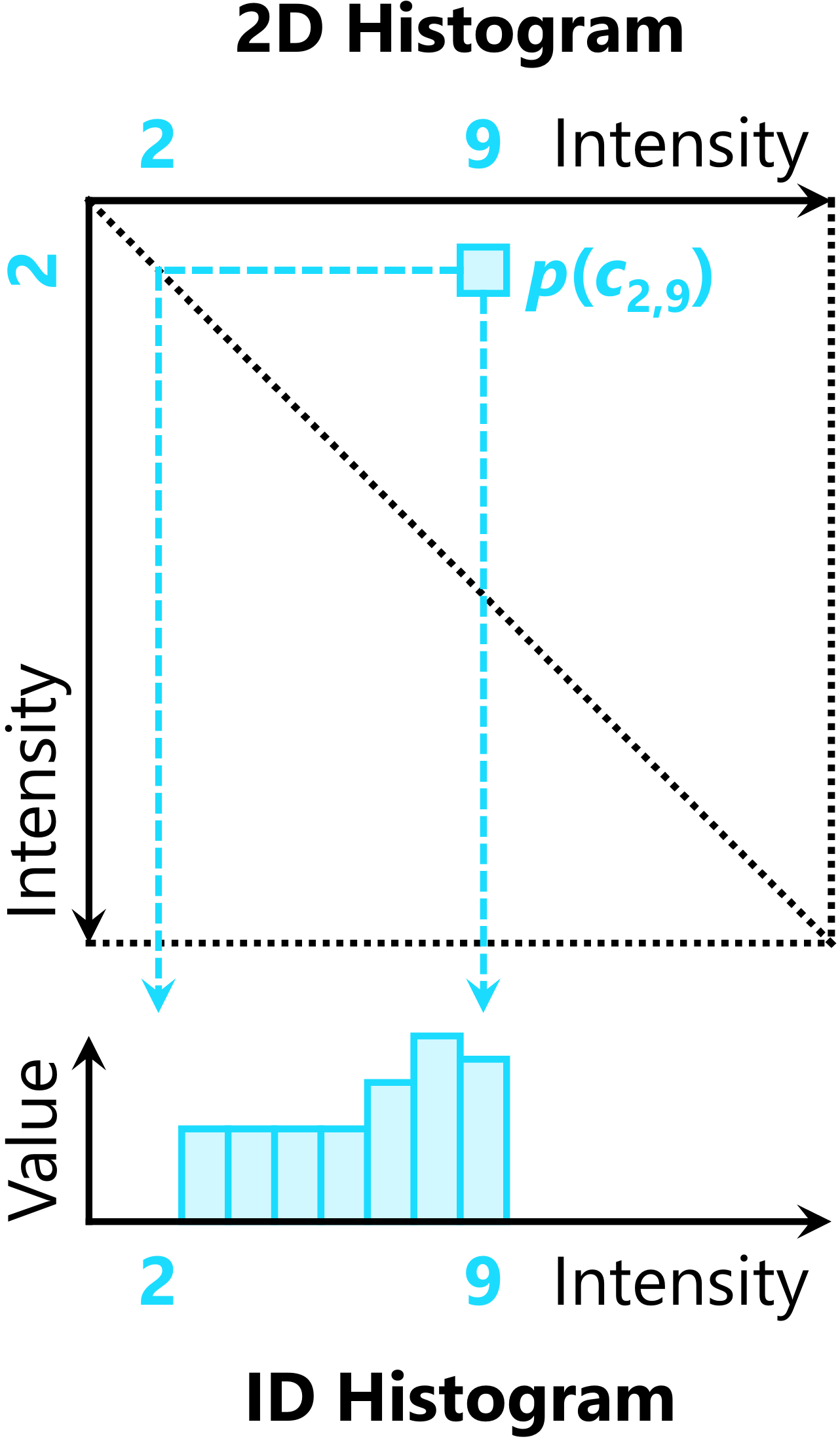}
\label{subf:rope}}
\caption{Toy examples showing how the 2D histogram value $p(c_{i,j})$ (blue square) is embedded in the 1D histogram. While previous approaches overemphasize the frequently co-occurring intensities $i$ and $j$, ROPE allows for more adequate and moderate contrast enhancement by distributing $p(c_{i,j})$ over all $k\in(i,j]$.}
\label{f:motivation}
\end{figure}

\begin{figure}[!t]
\centering
\subfloat[Input]{
\includegraphics[width=.31\linewidth]{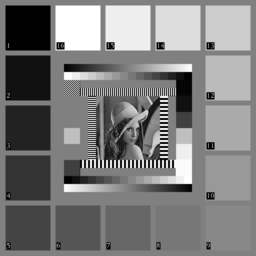}
\label{subf:testpat-input}}
\hfill
\subfloat[RG-CACHE~\cite{WuKK20}]{
\includegraphics[width=.31\linewidth]{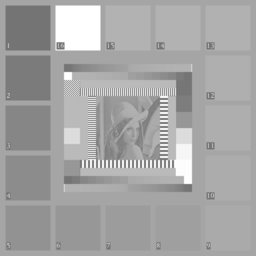}
\label{subf:testpat-rgcache}}
\hfill
\subfloat[ROPE]{
\includegraphics[width=.31\linewidth]{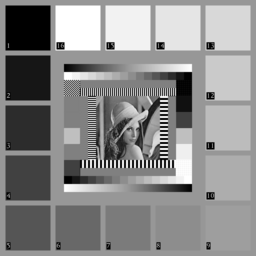}
\label{subf:testpat-rope}}
\\
\subfloat[Input]{
\includegraphics[width=.31\linewidth]{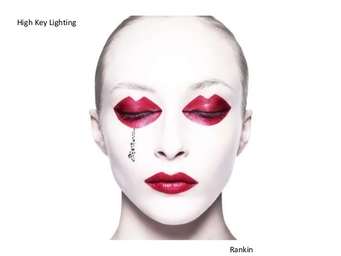}
\label{subf:a1-input}}
\hfill
\subfloat[RG-CACHE~\cite{WuKK20}]{
\includegraphics[width=.31\linewidth]{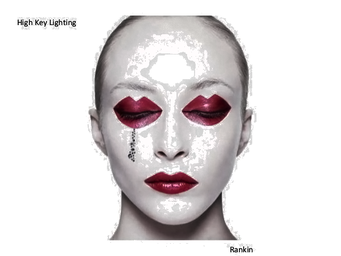}
\label{subf:a1-rgcache}}
\hfill
\subfloat[ROPE]{
\includegraphics[width=.31\linewidth]{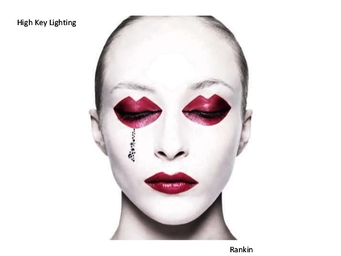}
\label{subf:a1-rope}}
\caption{Comparison of RG-CACHE~\cite{WuKK20} and ROPE.}
\label{f:rgcache}
\end{figure}

In CVC~\cite{CelikT11}, given two intensities $i$ and $j$, their 2D histogram value $p(c_{i,j})$ is voted into the bin of the larger intensity $j$ and added to the 1D histogram value $p(o_j)$, as illustrated in Fig.~\ref{subf:cvc}. Similarly, CACHE~\cite{WuLHK17,WuKK20} votes $p(c_{i,j})$ into the bins of both intensities (Fig.~\ref{subf:cache}). These schemes overemphasize $p(o_i)$ and/or $p(o_j)$ and thus tend to cause precipitous brightness fluctuation in very dark or bright image areas as shown in Figs.~\ref{subf:testpat-rgcache} and \ref{subf:a1-rgcache}. Instead, we aim to find a proper method to distribute $p(c_{i,j})$ over all $k\in(i,j]$ for more adequate contrast enhancement (Fig.~\ref{subf:rope} as well as Figs.~\ref{subf:testpat-rope} and \ref{subf:a1-rope}).

Our thinking is as follows. Wu et al.~\cite{WuLHK17,WuKK20} have revealed that in HE, for any $k\in(0,K)$, the degree of contrast enhancement (CE) between $k-1$ and $k$ is ultimately proportional to $p(o_k)$. Thus, for any $i,j\in[0,K)$, the degree of CE between $i$ and $j$ is proportional to $\sum_{k=i+1}^{j}p(o_k)$. Meanwhile, related studies~\cite{CelikT11,Celik12} suggested that in 2DHE, $p(c_{i,j})$ needs to be defined in such a way that it is positively correlated with the requirement of CE between $i$ and $j$. These two insights lead to the inference that $p(o_k)$ should be modeled such that $p(c_{i,j})\propto\sum_{k=i+1}^{j}p(o_k)$. This confirms our motivation above: $p(c_{i,j})$ should not be delegated to $p(i)$ and/or $p(j)$ alone, but should be distributed over all $k\in(i,j]$.

Now, we would like to build a bridge between $p(o_k)$ and $p(c_{i,j})$. We assume that $o_k$ and $c_{i,j}$ are dependent on each other. If we consider $o_k$ to be a marginal event, its distribution $p(o_k)$ can be obtained by marginalizing over $p(c_{i,j})$:
\begin{equation}
p(o_k)=\sum_{i=0}^{K-2}\sum_{j=i+1}^{K-1}p(o_k|c_{i,j})p(c_{i,j}),
\label{e:marginal}
\end{equation}
where $p(o_k|c_{i,j})$ is the conditional probability of $o_k$ given $c_{i,j}$. This formula implies that the 1D histogram value $p(o_k)$ can be modeled as a weighted average of all 2D histogram values $p(c_{i,j})$. The conditional probabilities $p(o_k|c_{i,j})$ act as weights.

Recall that it is necessary to determine $p(o_k)$ in such a way that $p(c_{i,j})\propto\sum_{k=i+1}^{j}p(o_k)$. Therefore, it is reasonable to assume that $o_k$ and $c_{i,j}$ are dependent on each other, i.e., $p(o_k|c_{i,j})\neq0$, if and only if $k\in(i,j]$; otherwise, $o_k$ and $c_{i,j}$ are mutually exclusive and $p(o_k|c_{i,j})=0$. In view of this, we introduce a significance factor $s_k$ for all $k\in[0,K)$ and define the conditional probabilities by
\begin{equation}
p(o_k|c_{i,j})=
\begin{cases}
\displaystyle \frac{s_k}{\sum_{k'=i+1}^{j}s_{k'}}&\text{if $k\in(i,j]$}\\
0&\text{otherwise}.
\end{cases}
\label{e:conditional}
\end{equation}

The intensity value $k$, which requires a greater contrast between $k-1$ and $k$, should have a greater value of $s_k$, and vice versa. However, we have no idea which intensity values are more important than others. In this study, we propose to determine $s_k$ through an iterative method. Let $t\in[1,\tau]$ be the index of the iteration and $\tau$ be the maximum number of iterations. When $t=1$, we initialize $s^{(1)}_k=\sfrac{1}{K}$ for all $k\in[0,K)$ and compute $p^{(1)}(o_k)$ using Eq.~\ref{e:marginal}. For all $t>1$, we update $s^{(t)}_k$ with $s^{(t)}_k=p^{(t-1)}(o_k)$ and recalculate $p^{(t)}(o_k)$ using the updated significance factor. In this way, we can guarantee that $o_k$ with a larger probability $p(o_k)$ tends to have a larger significance factor and thus tends to receive more contribution from the 2D histogram values $p(c_{i,j})$. Empirically, we found that two iterations are sufficient for ROPE to achieve satisfactory performance.

Fig.~\ref{f:rgcache} compares RG-CACHE~\cite{WuKK20} and our approach. Using RG-CACHE, dark and bright intensity values fluctuated sharply, causing blurred contrast (Fig.~\ref{subf:testpat-rgcache}) and eerie artifacts (Fig.~\ref{subf:a1-rgcache}). In comparison, ROPE achieved more high-quality and reliable enhancement.

\subsection{Embedding Reflectance in 2D Histogram}
\label{subs:rope}

Next, we describe how to construct the 2D histogram of the grayscale image $\mathbf{A}_\mathrm{in}$. In CVC~\cite{CelikT11}, the histogram value $p(c_{i,j})$ is determined as the co-occurrence frequency of the intensity value $j$ in the local window centered on the pixel of intensity $i$, further weighted by $|i-j|$. RG-CACHE~\cite{WuKK20} directly constructs a 1D histogram by incorporating the gradient of image reflectance into the density estimation. In this study, we borrow the idea of RG-CACHE to mitigate the negative effect of dark lighting conditions but embed the image reflectance into a 2D histogram instead of a 1D histogram.

\begin{figure}[!t]
\centering
\subfloat[Input $\mathbf{A}_\mathrm{in}$]{
\includegraphics[height=.31\linewidth]{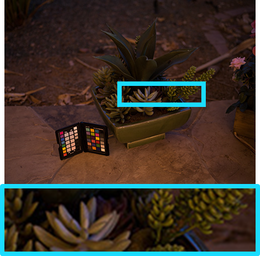}
\label{subf:3-input}}
\hfill
\subfloat[$p(c_{i,j})$ for CVC]{
\includegraphics[height=.31\linewidth]{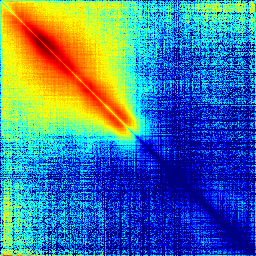}
\label{subf:3-2dh-cvc}}
\hfill
\subfloat[$p(c_{i,j})$ for ROPE]{
\includegraphics[height=.31\linewidth]{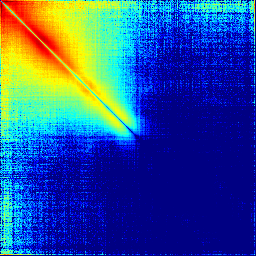}
\label{subf:3-2dh-rope}}
\\
\subfloat[1D histogram $p(o_k)$]{
\includegraphics[height=.31\linewidth]{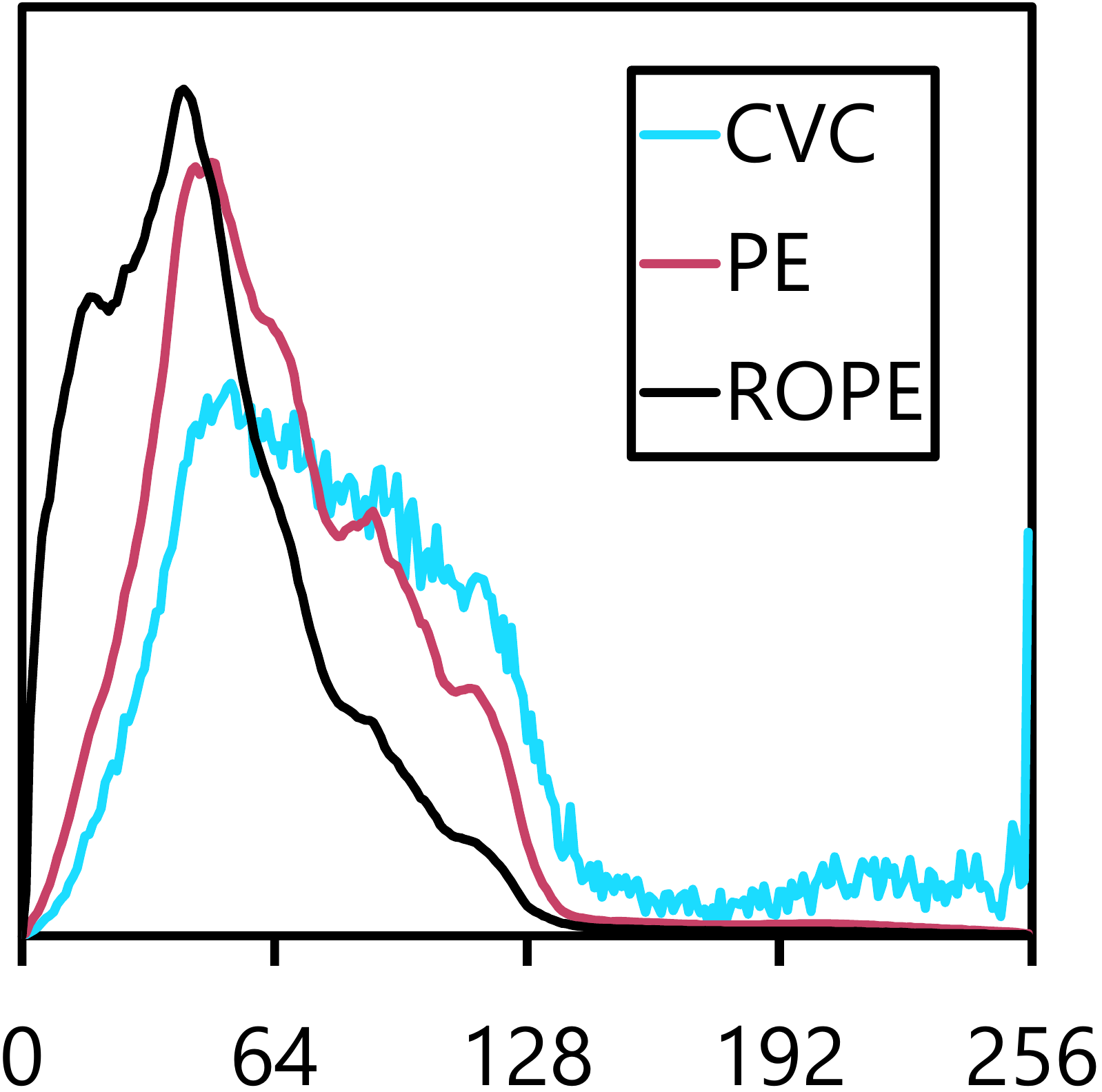}
\label{subf:3-1dh}}
\hfill
\subfloat[Mapping function $T$]{
\includegraphics[height=.31\linewidth]{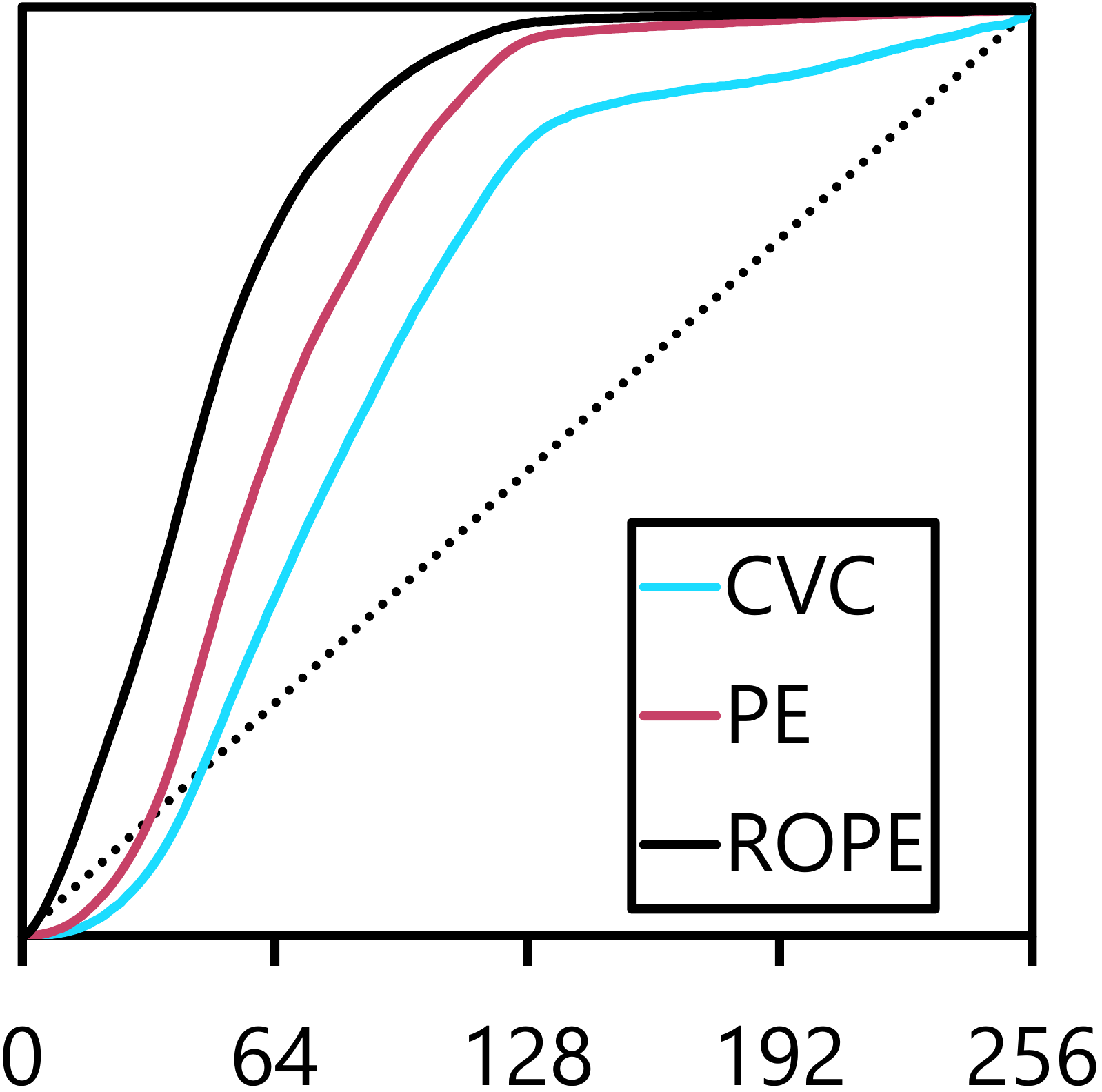}
\label{subf:3-mapping}}
\hfill
\subfloat[Reflectance $\mathbf{R}$]{
\includegraphics[height=.31\linewidth]{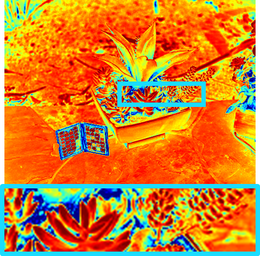}
\label{subf:3-reflectance}}
\\
\subfloat[$\mathbf{A}_\mathrm{out}$ for CVC]{
\includegraphics[height=.31\linewidth]{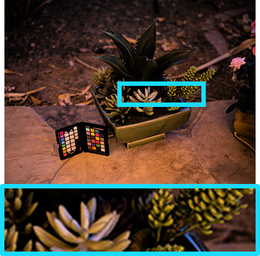}
\label{subf:3-cvc}}
\hfill
\subfloat[$\mathbf{A}_\mathrm{out}$ for PE]{
\includegraphics[height=.31\linewidth]{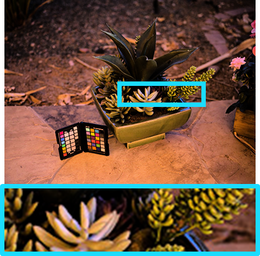}
\label{subf:3-pe}}
\hfill
\subfloat[$\mathbf{A}_\mathrm{out}$ for ROPE]{
\includegraphics[height=.31\linewidth]{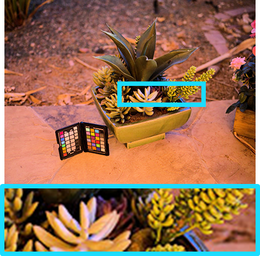}
\label{subf:3-rope}}
\caption{Comparison of ROPE with CVC~\cite{CelikT11} and PE. \protect\subref{subf:3-2dh-cvc} and \protect\subref{subf:3-2dh-rope}: 2D histograms whose $x$ and $y$-axes are intensities in $[0,256)$; histogram values are shown in color on a log scale. In \protect\subref{subf:3-pe}, PE means the 2D histogram of CVC was used for equalization.}
\label{f:lime-3}
\end{figure}

Let $\mathbf{I}$ and $\mathbf{R}$ be the illumination and the reflectance components of $\mathbf{A}_\mathrm{in}$, respectively. We first use the relative total variation (RTV) approach~\cite{XuYXJ12,GuoLL17} as an edge-preserving filter~\cite{XuLXJ11,He0T13,ZhangSXJ14,ZhangXJ14,ShibataTO16,ShenZXJ17} to estimate $\mathbf{I}$. We then consider a modified Retinex model as the formation of $\mathbf{A}_\mathrm{in}$: $\mathbf{A}_\mathrm{in}=\mathbf{I}\circ e^\mathbf{R}$ (where $e$ is the Euler number). Thus, the definition of the reflectance becomes
\begin{equation}
\mathbf{R}=\ln(\mathbf{A}_\mathrm{in}\oslash\mathbf{I}).
\label{e:reflectance}
\end{equation}
Eq.~\ref{e:reflectance} calculates $\mathbf{R}$ in the logarithmic domain. It reveals much richer objectness information hidden in the dark areas because logarithmic scaling magnifies the difference between small quantities, as shown in Figs.~\ref{subf:3-input} and \ref{subf:3-reflectance}.

Let $\mathbf{R}=\{r(q)\}$, where $r(q)$ is the reflectance value of the pixel $q\in\mathbf{A}_\mathrm{in}$. We calculate the 2D histogram values by
\begin{equation}
p(c_{i,j})=\frac{\sum_{q\in\mathbf{A}_\mathrm{in}}\sum_{q'\in\mathcal{N}(q)}|r(q)-r(q')|\delta_{a(q),i}\delta_{a(q'),j}}{\sum_{q\in\mathbf{A}_\mathrm{in}}\sum_{q'\in\mathcal{N}(q)}|r(q)-r(q')|},
\label{e:2dhistogram}
\end{equation}
where $\mathcal{N}(q)$ is a $7\times7$ window centered on $q$ according to CVC, and $\delta_{\cdot,\cdot}$ is the Kronecker delta. In this study, it is assumed that $p(c_{i,j})=p(c_{j,i})$ and $p(c_{i,i})=0$ for all $i,j\in[0,K)$.

Recall that $p(c_{i,j})$ needs to be defined in such a way that it is positively correlated with the requirement of CE between $i$ and $j$ (Section~\ref{subs:se}). Eq.~\ref{e:2dhistogram} satisfies this condition exactly; the 2D histogram values capture the local pixel value differences in reflectance and are sensitive to the presence of meaningful objects hidden in the dark. As shown in Fig.~\ref{subf:3-reflectance}, most large differences in reflectance are present in the foreground objects, e.g., the plants in the center and on the right and the ColorChecker. These objects obviously require greater brightness and contrast (high CE requirement) than the background. Conversely, the reflectance values of the background, where contrast is of less importance (low CE requirement), are rather smooth. Once the 2D histogram is constructed, it is substituted into Eq.~\ref{e:marginal} to iteratively calculate the 1D histogram. The input image is then enhanced by HE as described in Section~\ref{subs:preliminaries}.

\begin{table}[!t]
\centering
\caption{Quantitative assessment. For the first four metrics, higher statistics indicate better quality, while LOE is the opposite. The best statistics per column are shown in bold.}
\label{t:quantitative}
\small
\begin{tabular*}{\linewidth}{@{\extracolsep{\fill}}lccccr}
\toprule
Approach & DE & EME & PD & PCQI & LOE \\
\midrule
No Enhancement & 7.17 & 15.7 & 27.9 & 1.00 & \textbf{0} \\
\midrule
LIME~\cite{GuoLL17} & 7.08 & 13.0 & 27.6 & 0.88 & 156.5 \\
NPIE~\cite{WangL18} & 7.33 & 17.4 & 27.8 & 0.99 & 98.9 \\
LECARM~\cite{RenYLL19} & 7.11 & 12.2 & 25.0 & 0.90 & 299.9 \\
\midrule
KIND~\cite{ZhangZG19} & 7.02 & 11.7 & 22.5 & 0.87 & 155.4 \\
\midrule
CVC~\cite{CelikT11} & 7.49 & 23.5 & 32.3 & \textbf{1.06} & \textbf{0} \\
RG-CACHE~\cite{WuKK20} & \textbf{7.64} & 26.0 & 37.6 & 1.02 & \textbf{0} \\
$\star$ ROPE & 7.62 & \textbf{32.3} & \textbf{40.1} & 1.04 & \textbf{0} \\
\bottomrule
\end{tabular*}
\end{table}

Fig.~\ref{subf:3-input} shows an example of input image $\mathbf{A}_\mathrm{in}$. Figs.~\ref{subf:3-2dh-cvc}--\ref{subf:3-mapping} show the 2D/1D histograms and the intensity mapping function $T$ obtained using CVC, PE, and ROPE (PE indicates our approach proposed in Section~\ref{subs:se}, but with the 2D histogram of CVC used). Compared to CVC, ROPE has more emphasis on darker pixels, especially for important objects, thanks to the incorporation of reflectance. As shown in Fig.~\ref{subf:3-cvc}, CVC made no significant changes to the input, whereas PE improved the visibility of image detail, thereby demonstrating the effectiveness of Eq.~\ref{e:marginal}. By harnessing the reflectance effectively, ROPE further boosted brightness and contrast for the most satisfying image enhancement.

\section{Experiments}
\label{s:experiment}

In Section~\ref{subs:rope}, we compared ROPE to CVC~\cite{CelikT11} and RG-CACHE~\cite{WuKK20}. In this section, we mainly compare ROPE with four state-of-the-art approaches: LIME~\cite{GuoLL17}, NPIE~\cite{WangL18}, LECARM~\cite{RenYLL19}, and KIND~\cite{ZhangZG19}. The first three are based on the Retinex model; the last one is based on deep learning. All these approaches were evaluated using 578 images from four datasets: LIME~\cite{GuoLL17}, USC-SIPI~\cite{usc}, BSDS500~\cite{ArbelaezMFM11}, and VONIKAKIS~\cite{vonikakis}. The size of the local window used in ROPE was set to $7\times7$, and the maximum number of iterations $\tau$ was set to two.

\subsection{Qualitative Assessment}
\label{subs:qualitative}

\begin{figure*}[!t]
\centering
\subfloat[Input]{
\includegraphics[width=.15\linewidth]{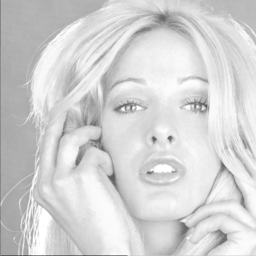}
\label{subf:4202-input}}
\hfill
\subfloat[LIME~\cite{GuoLL17}]{
\includegraphics[width=.15\linewidth]{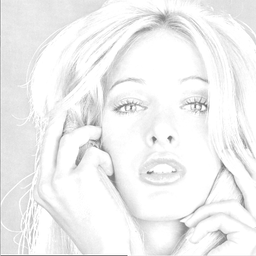}
\label{subf:4202-lime}}
\hfill
\subfloat[NPIE~\cite{WangL18}]{
\includegraphics[width=.15\linewidth]{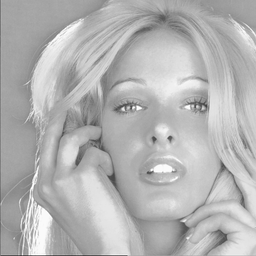}
\label{subf:4202-npie}}
\hfill
\subfloat[LECARM~\cite{RenYLL19}]{
\includegraphics[width=.15\linewidth]{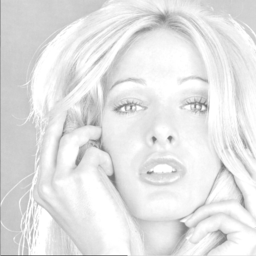}
\label{subf:4202-lecarm}}
\hfill
\subfloat[KIND~\cite{ZhangZG19}]{
\includegraphics[width=.15\linewidth]{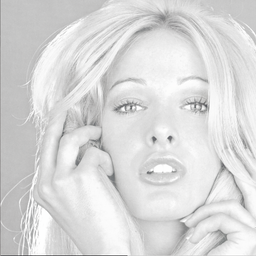}
\label{subf:4202-kind}}
\hfill
\subfloat[ROPE]{
\includegraphics[width=.15\linewidth]{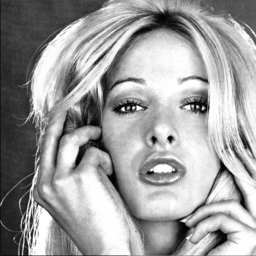}
\label{subf:4202-rope}}
\\
\subfloat[Input]{
\includegraphics[width=.15\linewidth]{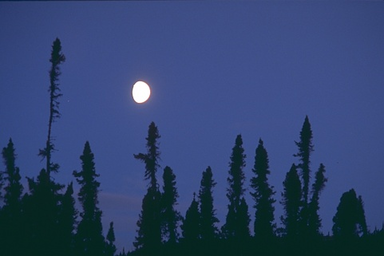}
\label{subf:238011-input}}
\hfill
\subfloat[LIME~\cite{GuoLL17}]{
\includegraphics[width=.15\linewidth]{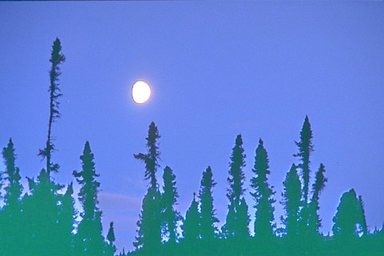}
\label{subf:238011-lime}}
\hfill
\subfloat[NPIE~\cite{WangL18}]{
\includegraphics[width=.15\linewidth]{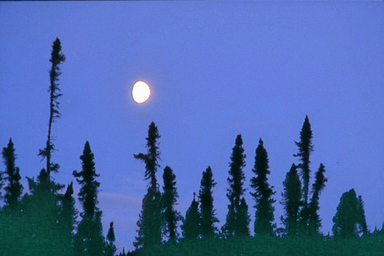}
\label{subf:238011-npie}}
\hfill
\subfloat[LECARM~\cite{RenYLL19}]{
\includegraphics[width=.15\linewidth]{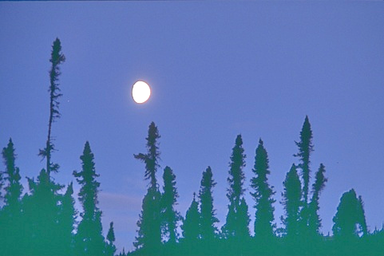}
\label{subf:238011-lecarm}}
\hfill
\subfloat[KIND~\cite{ZhangZG19}]{
\includegraphics[width=.15\linewidth]{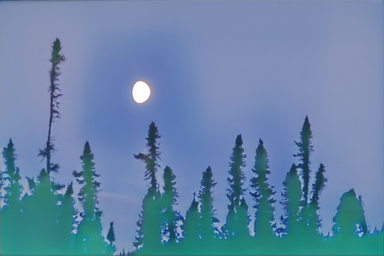}
\label{subf:238011-kind}}
\hfill
\subfloat[ROPE]{
\includegraphics[width=.15\linewidth]{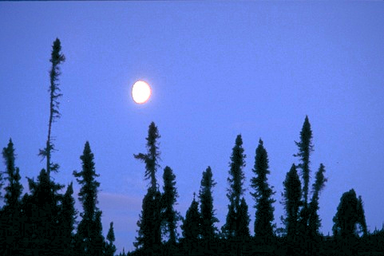}
\label{subf:238011-rope}}
\\
\subfloat[Input]{
\includegraphics[width=.15\linewidth]{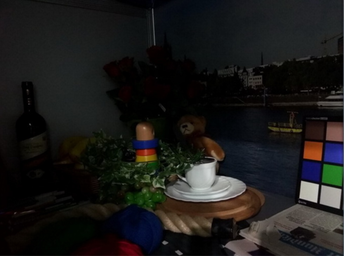}
\label{subf:8-input}}
\hfill
\subfloat[LIME~\cite{GuoLL17}]{
\includegraphics[width=.15\linewidth]{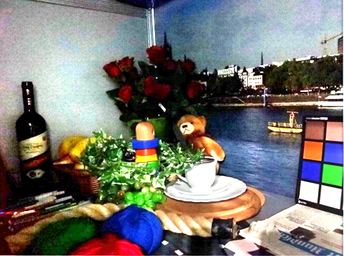}
\label{subf:8-lime}}
\hfill
\subfloat[NPIE~\cite{WangL18}]{
\includegraphics[width=.15\linewidth]{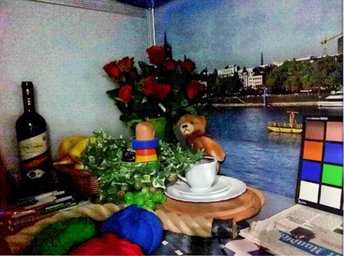}
\label{subf:8-npie}}
\hfill
\subfloat[LECARM~\cite{RenYLL19}]{
\includegraphics[width=.15\linewidth]{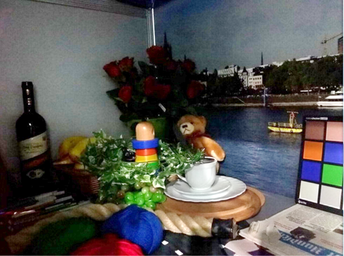}
\label{subf:8-lecarm}}
\hfill
\subfloat[KIND~\cite{ZhangZG19}]{
\includegraphics[width=.15\linewidth]{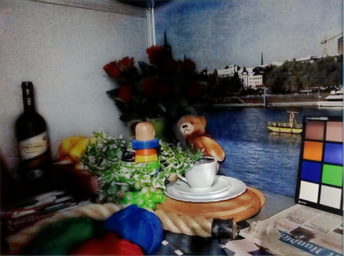}
\label{subf:8-kind}}
\hfill
\subfloat[ROPE]{
\includegraphics[width=.15\linewidth]{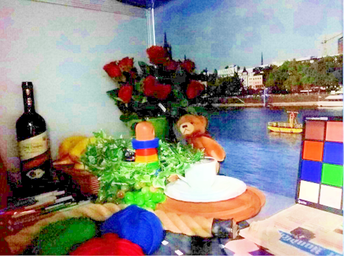}
\label{subf:8-rope}}
\\
\subfloat[Input]{
\includegraphics[width=.15\linewidth]{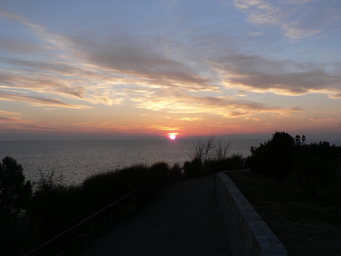}
\label{subf:P1010520-input}}
\hfill
\subfloat[LIME~\cite{GuoLL17}]{
\includegraphics[width=.15\linewidth]{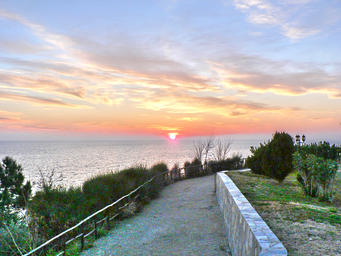}
\label{subf:P1010520-lime}}
\hfill
\subfloat[NPIE~\cite{WangL18}]{
\includegraphics[width=.15\linewidth]{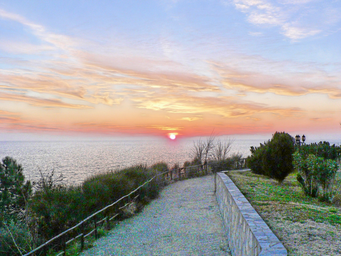}
\label{subf:P1010520-npie}}
\hfill
\subfloat[LECARM~\cite{RenYLL19}]{
\includegraphics[width=.15\linewidth]{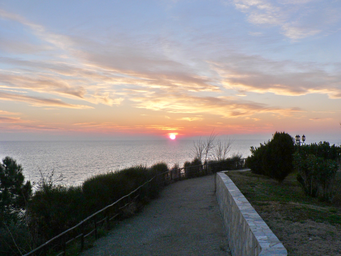}
\label{subf:P1010520-lecarm}}
\hfill
\subfloat[KIND~\cite{ZhangZG19}]{
\includegraphics[width=.15\linewidth]{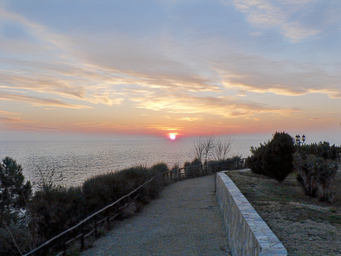}
\label{subf:P1010520-kind}}
\hfill
\subfloat[ROPE]{
\includegraphics[width=.15\linewidth]{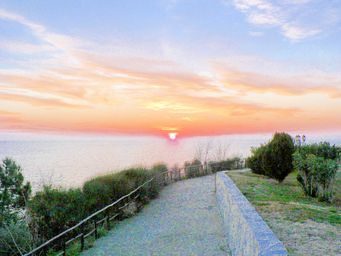}
\label{subf:P1010520-rope}}
\caption{Comparison of ROPE with state-of-the-art image enhancement (see supplemental material~\cite{appendix} for more examples). Images are from LIME~\cite{GuoLL17}, USC-SIPI~\cite{usc}, BSDS500~\cite{ArbelaezMFM11}, and VONIKAKIS~\cite{vonikakis}.}
\label{f:comparison}
\end{figure*}

The enhanced images obtained with the compared approaches are shown in Fig.~\ref{f:comparison}. KIND uses CNNs for image enhancement, but in a broader sense, it is also based on the Retinex model and so has similar performance to LIME, NPIE, and LECARM. Two common disadvantages of the four approaches are 1) a tendency to over-amplify brightness (first row) and 2) excessive saturation if color distortion was previously hidden in the dark areas of the input (second row). In comparison, ROPE does not suffer from these problems. The previous approaches essentially excel at enhancing low-light images, but there are exceptions, as shown in the third row. In this example, ROPE provided the most pleasant brightness, especially for the flowers in the center. All of these examples demonstrate the much greater adaptability and consistency of ROPE; our approach is capable of improving brightness sufficiently for dark images while avoiding excessive enhancement for normal-light images.

\subsection{Quantitative Assessment}
\label{subs:quantitative}

We objectively evaluated the image enhancement approaches using five metrics: discrete entropy (DE), EME, PD, and PCQI for contrast enhancement and LOE for naturalness. DE measures the amount of information in an image. EME~\cite{AgaianSP07} measures the average local contrast in an image. PD~\cite{ChenAPA06} measures the average intensity difference of all pixel pairs in an image. PCQI~\cite{WangMYWL15} measures the distortions of contrast strength and structure between input and output. LOE~\cite{WangZHL13} measures the difference in lightness order between the input and enhanced images. The lightness order means the relative order of the intensity values of two pixels. For DE, EME, PD, and PCQI, higher statistics indicate better quality, while LOE is the opposite.

Table~\ref{t:quantitative} shows the statistics averaged over 500 test images of BSDS500. Let us first focus on the contrast enhancement metrics. Since CVC, RG-CACHE, and ROPE are based on HE, they could maximize the range of intensity values and so achieved the highest scores. ROPE showed excellent contrast improvement capabilities, taking first place in EME and PD and second place in DE and PCQI. In comparison, none of the four approaches compared in Section~\ref{subs:qualitative} showed good performance.

In terms of image naturalness, all the approaches based on HE obtained the best LOE scores. These approaches have an inherent monotonicity constraint on the intensity mapping function $T$, so that the contrast enhancement does not change the order of intensity values in all pixels. In comparison, the approaches based on the Retinex model have poorer scores because they alter or eliminate the illumination component of the image, which results in a large variation in the lightness order.

\smallskip\noindent
\textbf{Computational Complexity.} Consider the processing of a grayscale image $\mathbf{A}_\mathrm{in}$ with $H\times{W}$ pixels and $K$ possible intensity values. The complexity of the reflectance estimation based on RTV~\cite{XuYXJ12,GuoLL17} is $\mathcal{O}(HW)$, which is the same as in LIME. The complexity of the 2D histogram construction (Eq.~\ref{e:2dhistogram}) is $\mathcal{O}({w^2}HW)$, where $w^2=7^2$ is the local window size. The 1D histogram construction (Eq.~\ref{e:marginal}) requires complexity $\mathcal{O}(K(K^2-1)/6)$. Although it appears to be computationally intensive, the processing time can be greatly reduced by exploiting convolutional operations. Applying the intensity mapping function $T$ to $\mathbf{A}_\mathrm{in}$ finally takes a complexity of $\mathcal{O}(HW)$. Given a color image, the complexities are the same as those described above, since ROPE is only applied to its intensity channel (Section~\ref{subs:preliminaries}).

\section{Conclusion}
\label{s:conclusion}

In this study, a novel image enhancement approach called ROPE is proposed. In this approach, an image is decomposed into illumination and reflectance components. The local pixel value differences in reflectance are embedded in a 2D histogram that captures the probability of intensity co-occurrence. ROPE derives a 1D histogram by marginalizing over the 2D histogram, assuming that intensity occurrence and co-occurrence are dependent on each other. Finally, an intensity mapping function is derived by HE for image enhancement. Evaluated on more than 500 images, ROPE surpassed state-of-the-art image enhancement approaches in both qualitative and quantitative terms. It was able to provide sufficient brightness improvement for low-light images while adaptively avoiding excessive enhancement for normal-light images.

\footnotesize
\bibliographystyle{IEEEbib}
\bibliography{refs}
\end{document}